\documentclass[letterpaper, 10 pt, conference]{ieeeconf}
\IEEEoverridecommandlockouts
\usepackage{graphics} 
\usepackage{epsfig} 

\usepackage{amsmath,amsthm} 
\usepackage{amssymb}  

\usepackage{hyperref}

\hypersetup{colorlinks,linkcolor={blue},citecolor={blue}} 

\usepackage{times} 
\usepackage[noadjust]{cite}

\usepackage{sjmacros}
\usepackage{thmstyles}
\usepackage{dsfont}
\usepackage{float}
\usepackage{multirow}
\usepackage{xcolor}
\usepackage{booktabs}
\usepackage{caption}
\usepackage{algorithm}
\usepackage{algpseudocode}

\title{\LARGE \bf
Heterogeneous Predictor-based Risk-Aware Planning\\
with Conformal Prediction in Dense, Uncertain Environments
}

\author{Jeongyong Yang$^{1*}$, KwangBin Lee$^{1*}$, and SooJean Han$^{1}$
\thanks{$^{*}$Both authors contributed equally to this work.}
\thanks{$^{1}$School of Electrical Engineering, Korea Advanced Institute of Science and Technology (KAIST), Daejeon 34141, Republic of Korea.
Correspondence to:
{\tt\small seiryu2238@kaist.ac.kr}}%
}

\begin{document}

\maketitle
\thispagestyle{empty}
\pagestyle{empty}

\begin{abstract}
Real-time planning among many uncertain, dynamic obstacles is challenging because predicting every agent with high fidelity is both unnecessary and computationally expensive. We present \emph{Heterogeneous Predictor-based Risk-Aware Planning} (H-PRAP), a framework that allocates prediction effort to where it matters. H-PRAP introduces the \emph{Probability-based Collision Risk Index} (P-CRI), a closed-form, horizon-level collision index obtained by calibrating a Gaussian surrogate with conformal prediction. P-CRI drives a router that assigns high-risk obstacles to accurate but expensive predictors and low-risk obstacles to lightweight predictors, while preserving distribution-free coverage across heterogeneous predictors through \emph{conformal prediction}. The selected predictions and their conformal radii are embedded in a chance-constrained model predictive control (MPC) problem, yielding receding-horizon policies with explicit safety margins. We analyze the safety-efficiency trade-off under prediction compute budget: more portion of low-fidelity predictions reduce residual risk from dropped obstacles, but in the same time induces larger conformal radii and degrades trajectory efficiency and shrinks MPC feasibility. Extensive numerical simulations in dense, uncertain environments validate that H-PRAP attains best balance between trajectory success rate (i.e., no collisions) and the time to reach the goal (i.e., trajectory efficiency) compared to single prediction architectures.
\end{abstract}

\section{Introduction}\label{sec:introduction}

Real-time planning in dynamic, uncertain environments remains a challenging problem, especially in crowded scenarios such as pedestrian-dense streets or industrial warehouses. 
Kinodynamic motion planning methods~\cite{karaman2010optimal, aoude_rrgp2013} are popular for integrating dynamics and constraints.
More importantly, optimization-based approaches like model predictive control (MPC)~\cite{ali_smpc2016, uncertainenv_mpc2023} provide a natural way to incorporate motion predictions, making them well-suited for real-time control.

Recent work has focused on improving safety by quantifying prediction uncertainty of moving obstacles through a distribution-free method, \emph{conformal prediction} (CP), and integrating this approach into MPC~\cite{conformal_prediction_planning,dixit2023adaptive}.
CP yields a \emph{conformal radius} around each predicted obstacle position that is guaranteed to contain the true position with a user-specified coverage level. 
However, many of these approaches rely on generating high-fidelity predictions for \emph{all} observable obstacles. This is not only inefficient but even computationally prohibitive in highly dense environments with a large number of moving obstacles. In practice, not all obstacles are equally significant to the agent's safety. Assigning computational resources uniformly to every obstacle neglects the fundamental \emph{accuracy--efficiency tradeoff}~\cite{liu2018} and creates an unnecessary bottleneck for real-time execution.

Therefore, we propose a \emph{heterogeneous} prediction framework that can integrate multiple predictors of varying accuracies and computational efforts, ranging from lightweight kinematic models (e.g., constant velocity) to more complex, data-driven approaches (e.g., LSTMs).
Inspired by System 1--System 2 thinking~\cite{kahneman2011thinking} and its engineering analogs~\cite{nakahira2021diversity}, this diversity of prediction models achieves a more favorable accuracy--efficiency balance than single-model approaches. 

To effectively leverage this framework, moving obstacles must be allocated to appropriate predictors based on their collision risk. 
Existing heuristic metrics like the collision risk index (CRI)~\cite{huang2020collisionCRI} rely on ad-hoc weightings and thresholds, providing no probabilistic guarantees and often misrepresenting the risk under high obstacle uncertainty.
To overcome this, we propose a novel confidence-driven metric, called \emph{Probability-based Collision Risk Index} (P-CRI). 
It delivers an analytic collision probability (non-central chi-squared CDF) under a Gaussian surrogate whose scale is calibrated with conformal prediction.

We integrate this risk-aware heterogeneous prediction into an MPC planner. At each planning step, a router driven by P-CRI routes high-risk obstacles to accurate but expensive predictors, routes low-risk obstacles to lightweight predictors, and passes predictor-specific conformal radii to the MPC constraints. This selective allocation enables safe and efficient real-time navigation in crowded settings while preserving distribution-free coverage for the obstacles considered by the planner. For practical safety-critical deployment, we also enforce a fixed per-cycle prediction compute budget: when the time limit is reached, H-PRAP prioritizes the highest-risk obstacles and skips or downgrades predictions for the rest.

In summary, the main contributions of this work are:
\begin{itemize}
    \item A \emph{heterogeneous prediction} framework that adaptively routes obstacles to predictors of different fidelities under a prediction compute budget, allocating computation where it most improves safety.
    \item P-CRI, a closed-form collision probability based risk index, computed as a non-central chi-squared CDF of a Gaussian surrogate whose variance is calibrated by conformal prediction.
    \item An MPC pipeline that uses predictor-specific conformal radii to enforce distribution-free safety; accompanying analysis provides coverage guarantees via conformal prediction for heterogeneous prediction.
    \item Analyses
    and experiments 
    on the trade-off between trajectory safety and efficiency under a fixed budget.
\end{itemize}


\section{Related Work}\label{sec:related_work}

Safe real-time navigation in crowded, stochastic environments commonly couples an optimizer--often Model Predictive Control (MPC)~\cite{park2021intention, alonso2018hierarchical}--with a motion predictor for surrounding obstacles. The performance of such a planner is fundamentally dependent on the fidelity and computational cost of those predictions.
Predictors span an accuracy--efficiency spectrum. Lightweight deterministic models, such as model-based predictors~\cite{berg2008reciprocal, van2011optimal}, integrate cleanly with MPC and support fast control, but they neglect the stochastic, socially compliant multi-modality observed in practical human or robot settings~\cite{rudenko2020human, williams2021recognizing}. Learning-based predictors~\cite{alahi2016social, gupta2018social} recover richer behavior but are expensive to evaluate online and thus scale only to a limited set of obstacles~\cite{okumura2022causal, rudenko2020survey}. This trade-off motivates the need for computation-aware prediction rather than relying on a single, uniform-fidelity approach.

\noindent\textbf{Navigation methods.} 
Reactive geometric controllers scale to many agents, but remains myopic and often encounter deadlocks in dense settings~\cite{van2011optimal,trautman2010unfreezing}. Interaction-aware approaches improve negotiation by coupling prediction and planning~\cite{chen2020rgl,sathyamoorthy2020frozone}, but incur nontrivial online cost and typically rely on heuristics to define interaction neighborhoods.

\noindent\textbf{Prediction pipelines.}
To capture multi-modality at scale, prediction pipelines extend learning-based predictors to many agents~\cite{alahi2016social,gupta2018social,salzmann2020trajectronpp}.
However, they often impose fixed radii or distance or time-to-collision pre-screens which are pragmatic, but brittle under high uncertainty~\cite{ahn2024sensors_ttc}. 
Therefore, to allocate high-fidelity prediction only to the most influential interactions while using lightweight models elsewhere, statistically calibrated uncertainty sets provide a principled criterion for \emph{which} neighbors/interactions {warrant large compute allocation} (i.e., high-fidelity prediction).

Among approaches that endow uncertainty sets with statistical validity, Conformal Prediction (CP) has been increasingly used across modern machine learning and is becoming a standard tool~\cite{zhou2025conformal,angelopoulos2021gentle}. Under exchangeability of data, split-conformal procedures provide finite-sample, distribution-free coverage for prediction errors~\cite{lei2018distribution,angelopoulos2024theoretical}. In planning and control, CP has been integrated with MPC to enforce safety by converting learned predictors into calibrated uncertainty radii with empirical coverage guarantees~\cite{conformal_prediction_planning,dixit2023adaptive}. This line of work underscores a broader shift toward principled uncertainty quantification in crowded, stochastic environments.

\begin{figure*}[t]
    \centering
    \includegraphics[width=0.8\linewidth]{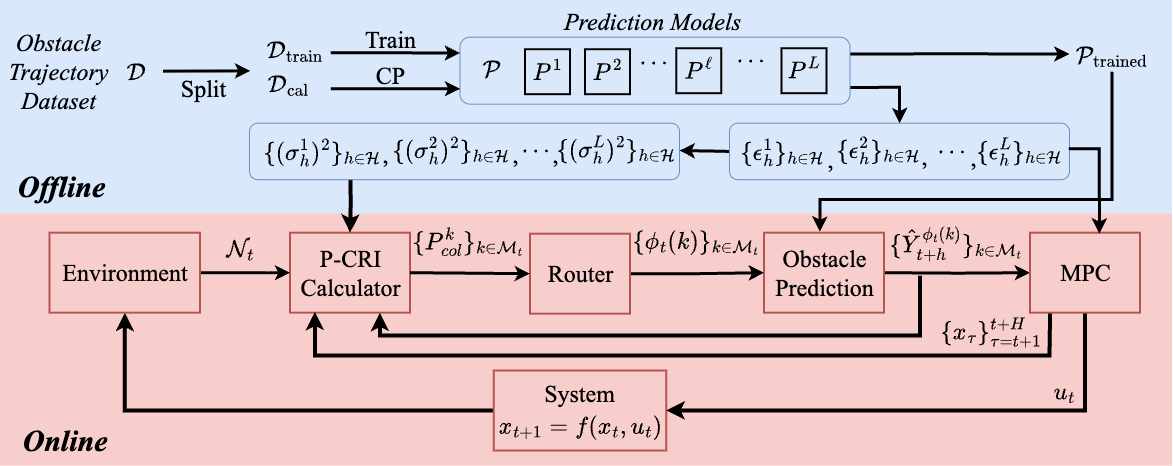}
    \caption{Overall H-PRAP architecture.
    \textbf{[Offline]} An obstacle trajectory dataset $\mathcal{D}$ is split into a training set $\mathcal{D}_{\text{train}}$ and a calibration set $\mathcal{D}_{\text{cal}}$. $\mathcal{D}_{\text{train}}$ is used to train the predictor set $\mathcal{P}$, while $\mathcal{D}_{\text{cal}}$ is used to pre-compute predictor-wise conformal radii $\{\epsilon^{\ell}_{h}\}_{h\in\mathcal{H}}$. These radii calibrate the Gaussian surrogate variances $\{(\sigma^\ell_{h})^{2}\}_{h \in \mathcal{H}}$ used in the risk metric, P-CRI.
    \textbf{[Online]} At each step, sensed obstacles are scored by the P-CRI Calculator; the Router assigns a predictor $\phi_t(k)$ to the sensed obstacles.
    The Obstacle Prediction module outputs future trajectories $\hat{Y}^{\phi_t(k)}_{t+h}$, which together with the corresponding radii feed the MPC to compute a safe control input $u_t$.}
    \label{fig:H-PRAP}
    \vspace{-0.4cm}
\end{figure*}

\section{Problem Formulation \& Preliminaries} \label{ProblemFormulation}
We consider scenarios in which an agent navigates toward a designated goal, avoiding dynamic obstacles.
The agent abides by some general discrete-time dynamics $\xvect_{t+1}{\,=\,}f(\xvect_t, \uvect_t)$ at each time $t{\,\in\,}\Zbb_{\geq 0}$, with state $\xvect_t{\,\in\,}\Rbb^n$, initial condition $\xvect_0{\,\triangleq\,}\zeta$ and control input $\uvect_t{\,\in\,}\Rbb^m$. 
The agent initially knows a static map of the environment and the target position.
The agent has sensors that observe all local changes in its sensing range.

The dynamic environment contains a set $\Ncal{\,\triangleq\,}\{1,\cdots,N\}$ ($N{\,\in\,}\Nbb$) of moving obstacles.
We define $\Ncal_t{\,\subseteq\,}\Ncal$ ($N_t{\,\triangleq\,}\abs{\Ncal_t}$) as the set (number) of obstacles within the agent's sensing range at time $t$.
Instead of predicting trajectories of all observed obstacles, a subset $\Mcal_t{\,\subseteq\,}\Ncal_t$ is selected for prediction.

To model the behavior of moving obstacles, let $\mathcal{D}$ denote the probability distribution over obstacle trajectories. Let $ (Y_0, Y_1, \ldots){\,\sim\,}\mathcal{D} $ denote a trajectory for a single moving obstacle, where $Y_t \in \mathbb{R}^n$ is an obstacle state at time $t$.
\begin{assumption}[Obstacles Non-interacting with Agent]\label{assum:noninteraction}
    The distribution $\Dcal$ does not depend on the control inputs $\uvect_t$ or the agent trajectory $\xvect_t$ for all $t{\,\geq\,}0$. In other words, the agent does not influence the future behavior of the obstacles: $\left(Y_{t+1}, Y_{t+2}, \ldots\right)\perp\!\!\!\perp\left(\xvect_0, \ldots, \xvect_t\right) | \left(Y_0, \ldots, Y_t\right)$.
\end{assumption}
\noindent The assumption, commonly used in robot planning (e.g.,~\cite{conformal_prediction_planning}), prevents distribution shift and retains conformal coverage. 
\begin{assumption}[i.i.d. Data]\label{assum:iid}
We assume access to a dataset $ D {\,\triangleq\,} \{Y^{(1)}, \ldots, Y^{(I)}\} $, where each $Y^{(i)}  {\,=\,}  (Y^{(i)}_0, Y^{(i)}_1, \ldots) $ represents a complete trajectory data of $i$-th obstacle
over time. 
Each $Y^{(i)}$ is drawn independently from the same
distribution $\mathcal{D}$ (i.e., $Y^{(i)}\sim\mathcal{D})$.\footnote{This can be relaxed to exchangeable trajectory data for the conformal prediction in Sec.~\ref{subsec:heterogeneous_conformal}; i.i.d. is a sufficient (stronger) condition.}
\end{assumption}

The dataset $D$ is partitioned into a training data set $D_{\text{train}}$ and a calibration dataset $D_{\text{cal}}$ (i.e., $D_{\text{train}}{\,\cap\,} D_{\text{cal}} {\,=\,} \varnothing$) to train prediction models and obtain the conformal radii, respectively.
At time $t$, a predictor $P$ trained from $D_{\text{train}}$ produces multi-step forecasts
$\hat Y_{t+h}$
for $h {\,\in\,} \mathcal{H}$, from the previous trajectory $Y_{0:t}$, where $\mathcal{H} {\,\triangleq\,} \{1,\ldots,H\}$ with prediction horizon $H\in\Nbb$. For each trajectory $Y^{(i)}$ sampled from $D_{\text{cal}}$, the predictor is run on the prefix $Y^{(i)}_{0:t}$ to obtain $\hat Y^{(i)}_{t+h}$ and compute the nonconformity score $R^{(i)}_{h} {\,\triangleq\,} \big\|Y^{(i)}_{t+h}-\hat Y^{(i)}_{t+h}\big\|$.
Letting $R^{(1)}_{h} {\,\le\,} \cdots {\,\le\,} R^{(N)}_{h}$ be sorted,
we set the confidence level as $1-\bar\delta$ (failure probability set as $\overline{\delta})$, and define the conformal radius as $\epsilon_{h} {\,\triangleq\,} R^{(q)}_{h}$ where $q {\,=\,} \Big\lceil \! (N{+}1)\,(1-\bar\delta) \! \Big\rceil$ with $R^{(N{+}1)}_{h}{\,=\,}+\infty$.

\begin{lemma}\label{lem:cp_single_step_coverage}
Given a trajectory $Y_{t+h} {\,\sim\,} \mathcal{D}$, and predictions $\hat Y_{t+h}$, the calibration dataset $D_\text{cal}$, and a user-chosen confidence level $1-\overline\delta \in (0,1)$, we have
\begin{equation}\label{eq:lemma_1}
\mathbb{P}\!\left(\big\|Y_{t+h}-\hat Y_{t+h}\big\|\le \epsilon_{h}\right)\ \ge\ 1-\bar\delta,\quad\forall h {\,\in\,} \mathcal{H}
\end{equation}
\end{lemma}
\begin{lemma}\label{lem:cp_horizontal_coverage}
Under the same assumption in~\lem{cp_single_step_coverage}, let $\overline{\delta} = \delta / H$. We have
\begin{equation}\label{eq:het_conformal_pred_prob_horz_intersection}
    \Pbb\Biggl(\,\bigcap_{h\in\mathcal{H}} 
    \Bigl\{\|Y_{t+h} - \hat{Y}_{t+h}\|\le \epsilon_{h}\Bigr\}\Biggr) \ge 1 - \delta.
\end{equation}
\end{lemma}
\noindent
See Lemma~\ref{lem:cp_single_step_coverage} and Lemma~\ref{lem:cp_horizontal_coverage} in~\cite{conformal_prediction_planning} for the proofs.


\section{Safe Planning with Heterogeneous Prediction}\label{sec:framework}

Most trajectory planners use a single \emph{homogeneous} predictor for all obstacles. We instead adopt a \emph{heterogeneous} scheme:  obstacles are routed to different predictors, with various accuracies and computation efforts while preserving reliable uncertainty bounds.
We call the framework \emph{heterogeneous Predictor-based Risk Aware Planning} (H-PRAP). Fig.~\ref{fig:H-PRAP} provides an overview.

Let $\Pcal{\,=\,}\{P^1,\ldots,P^L\}$ be the available predictors with index set $\{1,\ldots,L\}$. Each predictor $P^{\ell}$ is trained with the training dataset $D_{\text{train}}$.
We precompute \emph{per-predictor conformal radii} $\{\epsilon_{h}^\ell\}_{h\in\mathcal{H}}$ offline using only the calibration dataset $D_{\text{cal}}$
(Sec.~\ref{subsec:heterogeneous_conformal}).

At each time $t$, every observed obstacle in $\Ncal_t$ is labeled by a routing function $
\phi_t: \Ncal_t \to \{0,1,\ldots,L\}$.
Internally, $\phi_t$ computes probability-based collision risk index for each obstacle $k$ and assigns a label $\ell$ (Secs.~\ref{subsec:pcri_index}, \ref{subsec:routing}). Label $\ell=0$ denotes no prediction, while $\ell\ge 1$ routes $k$ to predictor $P^\ell{\,\in\,}\Pcal$. The predicted sets are thus $
\Mcal_t{\,\triangleq\,}\{k\in\Ncal_t: \phi_t(k)\neq 0\}$, $ 
\Mcal_t^\ell \triangleq \{k\in\Ncal_t: \phi_t(k)=\ell\}$ where $ \ell\in\{1,\ldots,L\}$ (and cardinalities $M_t\triangleq|\mathcal{M}_t|, \ M_t^\ell \triangleq |\mathcal{M}_t^\ell|)$.

Finally, we use a chance-constrained MPC that utilizes the predicted trajectories and their conformal radii.
At each step, the controller chooses control inputs that keep the agent outside the circular regions formed by conformal radii with probability at least $1-\delta$, while driving towards the goal.
This yields a receding-horizon control with distribution-free probabilistic safety margins (Sec.~\ref{subsec:mpc_optimization}).

\subsection{Conformal Prediction for Heterogeneous Prediction}\label{subsec:heterogeneous_conformal}

To leverage conformal prediction in a heterogeneous predictor setting, we formulate a probabilistic bound for multiple different predictors. 
For each obstacle $k{\,\in\,}\Mcal_t$ at time $t$ using prediction model $P^{\phi_{t}(k)}{\,\in\,}\mathcal{P}$, we want to ensure
\begin{equation}\label{eq:het_conformal_pred_prob}
    \Pbb\left(\lVert Y^{k}_{t+h} - \hat{Y}^{\phi_{t}(k)}_{t+h}\rVert \leq \epsilon^{\phi_{t}(k)}_{h}\right) \geq 1-\overline{\delta},\quad\forall h {\,\in\,} \mathcal{H}.
\end{equation}
For obstacle $k$, $Y_{t+h}^{k}$ is the ground truth trajectory, $\hat{Y}_{t+h}^{\phi_{t}(k)}$ is the predicted trajectory with its allocated prediction model $P^{\phi_{t}(k)}$, and
$\epsilon^{\phi_{t}(k)}_{h}$ denotes the per-predictor conformal radii when routed to $P^{\phi_{t}(k)}$.
\footnote{The superscript of $Y$ denotes the index of obstacle (e.g., $k$), but that of $\hat{Y}$ denotes predictor index (e.g., $\ell\in\{1,...,L\}$, where $\ell {\,=\,} \phi_t(k)$).}
Essentially, \eqref{eq:het_conformal_pred_prob} says the ground-truth position of the obstacle lies within a ball of radius $\epsilon^{\phi_{t}(k)}_{h}$ centered at $\hat{Y}_{t+h}^{\phi_{t}(k)}$ with probability no less than $1-\overline{\delta}$.
Note that by choosing the confidence level $1-\overline{\delta}$, the user can explicitly control uncertainty bounds.~\eqref{eq:het_conformal_pred_prob} can be extended to the entire prediction horizon by choosing $\overline{\delta} \triangleq \delta/H$:
\begin{equation}
    \Pbb\Bigl(\,\bigcap_{h\in\mathcal{H}}
    \{\|Y^{k}_{t+h} - \hat{Y}^{\phi_{t}(k)}_{t+h}\|\le \epsilon^{\phi_{t}(k)}_{h}\}\Bigr) \;\ge\; 1 - \delta.
    \label{eq:het_conformal_pred_prob_horz}
\end{equation}

\subsection {Probability-based Collision Risk Index (P-CRI)}\label{subsec:pcri_index}
To allocate computation resources effectively, we estimate the collision risk of each obstacle over the prediction horizon, using the trajectory predicted in the previous step. We introduce a \emph{Gaussian surrogate} to enable closed-form risk computation for predictor $P^{\ell}$:
\begin{equation}
    \tilde{Y}_{t+h-1}^{\ell} {\,\sim\,} \mathcal{N}\!\left( \hat{Y}^{\ell}_{t+h-1}, (\sigma^{\ell}_{h})^2 I \right),\quad\forall h {\,\in\,} \mathcal{H}.
\end{equation}
The covariance $(\sigma^{\ell}_{h})^2 I$ is chosen offline so that $1{\,-\,}\overline{\delta}$ confidence level of this Gaussian matches the precomputed conformal radius.
Here, $\tilde{Y}_{t+h-1}^{\ell}$ is a tractable approximation of the true (possibly non-Gaussian) distribution.
%
Since the squared Mahalanobis distance of a multivariate Gaussian follows a chi-squared distribution, the corresponding cumulative distribution function (CDF) satisfies
\begin{equation}\label{eq:chi_squared_confidence}
    \mathbb{P}\left( \frac{\| \tilde{Y}^{\ell}_{t+h-1} - \hat{Y}^{\ell}_{t+h-1} \|^2}{ (\sigma^{\ell}_{h})^2} \leq \chi^2_n(1 - \overline{\delta}) \right)= 1 - \overline{\delta},
\end{equation}
where $\chi^2_n(1 - \overline{\delta})$ denotes the $1 - \overline{\delta}$ quantile of the chi-squared distribution.
By combining~\eqref{eq:lemma_1} and~\eqref{eq:chi_squared_confidence} ($\epsilon_h^\ell$ is for predictor $P^\ell$), the variance can be precomputed to 
\begin{equation} \label{eq:variance_kai}
    (\sigma^{\ell}_{h})^2 = (\epsilon_h^{\ell})^2 / \chi^2_n(1 - \overline{\delta}).
\end{equation}

After $\{(\sigma^{\ell}_{h})^2\}_{h \in \Hcal}$ obtained offline for all predictors $P^\ell$, the following process is done online.
Given the agent's predicted position $\xvect_{t+h-1}$ (Sec.~\ref{subsec:mpc_optimization}), the relative position vector $\mathbf{\Delta x}_{t+h-1}^{k}{\,\triangleq\,}  \tilde{Y}^{\ell}_{t+h-1}  {\,-\,}  \xvect_{t+h-1}$ is also Gaussian with mean $\hat{Y}^{\ell}_{t+h-1} {\,-\,} \xvect_{t+h-1}$ and covariance $(\sigma^{\ell}_{h})^2 I$, where $\ell {\,=\,} \phi_{t-1}(k)$. 
The obstacle $k$ is considered a collision risk if its distance to the agent is less than a distance $r_\text{margin}$.
Since $\|\mathbf{\Delta x}_{t+h-1}^{k}\|^2$ is the square of a nonzero-mean Gaussian vector, it also follows a non-central chi-squared distribution, and the per-step collision probability of obstacle $k$ at $h {\,\in\,} \mathcal{H}$ is
\begin{equation}\label{eq:collision_probability_chi_squared}
    p^{k}_{t+h-1}
    {=\,} \mathbb{P} \! \left( \| \mathbf{\Delta x}_{t+h-1}^{k} \|^2 \leq r_\text{margin}^2 \right)\!
    {=} F_{\chi^2_n(\lambda_h)} \! \left( \frac{r_\text{margin}^2}{(\sigma^{\ell}_{h})^2} \right) \!,
\end{equation}
where $F_{\chi^2_n(\lambda)}$ is the CDF of the non-central chi-squared distribution with non-centrality parameter $\lambda_h {\,\triangleq\,} \| \hat{Y}^{\ell}_{t+h-1} {\,-\,} \xvect_{t+h-1} \|^2 / (\sigma^{\ell}_{h})^2 $.

Finally, we aggregate the probabilities into a collision risk index $P_{t,\text{col}}^k$ through a monotone function $f$:
\begin{equation}\label{eq:risk_aggregation_general}
    P_{t,\text{col}}^k = f\!\left(\{p_{t+h-1}^k\}_{h\in\mathcal{H}}\right).
\end{equation}
A simple choice is $P_{t,\text{col}}^k{\,=\,}\sum_{h\in\mathcal{H}} p_{t+h-1}^k$, which corresponds to Boole’s inequality on the 
probability of at least one collision event within the horizon.

\subsection{Routing Obstacles to Prediction Models.}\label{subsec:routing}
At each time $t$, every observed obstacle $k {\,\in\,} \Mcal_t$ is routed to a prediction model according to its collision risk index through $\phi_t(k)$.
We implement a simple thresholding mechanism to route. Let $\{ \theta_0, \theta_1, \ldots, \theta_L \}$ be a sequence of fixed thresholds with $\theta_0 {\,=\,} 0$ and $\theta_L {\,=\,} \infty$. 
If $P^{k}_{t,\text{col}} {\,\in\,} [\theta_{\ell}, \theta_{\ell+1})$, then the routing function assigns $\phi_{t}(k) {\,=\,} \ell$, i.e., obstacle $k$ is predicted using model $P^{\ell}$.

Thus, high-risk obstacles are assigned to accurate but expensive predictors, while low-risk ones use lightweight predictors. Obstacles with negligible risk ($\phi_t(k){\,=\,}0$) are assigned to a trivial motion model (e.g., constant velocity) only to keep~\eqref{eq:collision_probability_chi_squared} well defined; such obstacles are excluded from the MPC constraints since they are not safety-critical.

\subsection{Model Predictive Control for Safe Planning }\label{subsec:mpc_optimization}

The MPC planner generates control inputs with collision avoidance constraints based on obstacle trajectory predictions at each time step $t$ as follows:
\begin{subequations}\label{eq:mpc_optimization}
\begin{alignat}{2}
&\min_{\uvect_{t:t+T-1}} \sum_{\tau=0}^{T-1} (\xvect_{t+\tau} - \xvect_g)^{\top} Q (\xvect_{t+\tau} - \xvect_g) + \uvect_{t+\tau}^{\top} R \uvect_{t+\tau} \notag\\
&\mkern100mu + (\xvect_{t+T} - \xvect_g)^{\top} Q_f (\xvect_{t+T} - \xvect_g) \\
\text{s.t.}\notag\\
&\mkern5mu\forall k \in \mathcal{M}_t, \uvect_{t+\tau} \in \mathcal{U},\;\xvect_{t+\tau+1} \in \mathcal{X}, 
\label{eq:mpc_constraints}\\
&\mkern5mu \xvect_{t+\tau+1} = f(\xvect_{t+\tau}, \uvect_{t+\tau}), 
   &&\mkern-200mu\forall \tau\in\{0,\dots,T-1\}
\label{eq:mpc_dynamics}\\
&\mkern5muc(\xvect_{t+h}, \hat{Y}_{t+h}^{\phi_{t}(k)}) 
   \geq K\,\epsilon_{h}^{\phi_{t}(k)}, 
   &&\mkern-200mu\forall h\in\{1,\dots,H\}.
\label{eq:mpc_safety_runningEx}
\end{alignat}
\end{subequations}
Here, $Q$, $Q_f$, and $R$ are symmetric, positive-definite cost matrices chosen by the user.
The MPC planning horizon is $T$, which is chosen to be greater than the obstacle prediction horizon $H$.
The states $\xvect_{t+\tau}$ and goal $\xvect_g$ are in the state space $\mathcal{X}\subseteq\Rbb^n$, and the control input $\uvect_{t+\tau}$ lies in the admissible control input set $\mathcal{U}\subseteq\Rbb^m$. We incorporate the conformal radius from Sec.~\ref{subsec:heterogeneous_conformal} into the safety constraint~\eqref{eq:mpc_safety_runningEx}, where $c:\mathbb{R}^n \times \mathbb{R}^n \rightarrow \mathbb{R}$ is a collision-avoidance function which is assumed $K$-Lipschitz in its second argument. A typical choice is $c(\xvect, \yvect)\triangleq\|\xvect-\yvect\|_2 - r_{\text{margin}}$.

We now establish a safety guarantee for every obstacle considered by the planner. The following results reinterpret the constraint in~\eqref{eq:mpc_safety_runningEx} as a chance-constraint.
\begin{proposition}[Following~\cite{conformal_prediction_planning}]\label{prop:mpc_total_constraint_bound}
Assume that:
(i) the collision-avoidance function $c$ is $K$-Lipschitz in its second argument,
(ii) MPC problem \eqref{eq:mpc_optimization} is feasible, and
(iii) conformal coverage~\eqref{eq:het_conformal_pred_prob_horz} holds over the prediction horizon $\Hcal$.
Then any feasible solution of \eqref{eq:mpc_optimization} satisfies
\begin{equation}
\mathbb P\Biggl(\,\bigcap_{h \in \mathcal{H}}
\{\,c(\xvect_{t+h}, Y^k_{t+h})\ge 0\,\}\Biggr) \;\ge\; 1-\delta.
\end{equation}
\end{proposition}

\begin{proof}
Let $\mathcal E_t$ denote the probability event in \eqref{eq:het_conformal_pred_prob_horz}. In $\mathcal E_{t}$, for every $h {\,\in\,} \mathcal{H}$,
the Lipschitz property in the second argument yields $c(\xvect_{t+h}, Y^k_{t+h}) {\,\ge\,} c(\xvect_{t+h}, \hat Y^{\phi_t(k)}_{t+h}) - K\|Y^k_{t+h}-\hat Y^{\phi_t(k)}_{t+h}\|$.
By the MPC safety constraint \eqref{eq:mpc_safety_runningEx}, we also have
$c(\xvect_{t+h}, \hat Y^{\phi_t(k)}_{t+h}) {\,\ge\,} K \epsilon^{\phi_t(k)}_{h}$.
Combining the two inequalities and using $\|Y^k_{t+h}-\hat Y^{\phi_t(k)}_{t+h}\| {\,\le\,} \epsilon^{\phi_t(k)}_{h}$ on $\mathcal E_t$ gives
\begin{equation*}
c(\xvect_{t+h}, Y^k_{t+h}) \ge K\Big(\epsilon^{\phi_t(k)}_{h}-\|Y^k_{t+h}-\hat Y^{\phi_t(k)}_{t+h}\|\Big) \ge 0.
\end{equation*}
Since $c(\xvect_{t+h}, Y^k_{t+h}) {\,\ge\,} 0$ holds for all $k {\,\in\,} \Mcal_t$ and all $h {\,\in\,\mathcal{H}}$ on $\mathcal E_t$, the proof is complete.
\end{proof}
\noindent Building upon this result, we extend the guarantee to ensure simultaneous safety for all obstacles considered by the planner. This is achieved by applying a Bonferroni correction to the individual failure probabilities.
\begin{corollary}\label{lem:mpc_total_constraint_bound}
For time $t$, let $\Mcal_t$ be the set of obstacles routed to predictors (i.e., $\phi_t(k){\,\neq\,} 0$).
Assume that:
(i) the collision-avoidance function $c$ is $K$-Lipschitz in its second argument,
(ii) MPC problem \eqref{eq:mpc_optimization} is feasible, and
(iii) the heterogeneous conformal coverage holds \emph{simultaneously} over the prediction horizon for all routed obstacles, namely:
\begin{equation}\label{eq:scene_cov_event_heterogeneous}
\mathbb P\Biggl(\,
\bigcap_{k\in \Mcal_t}\ \bigcap_{h\in\mathcal{H}}
\bigl\{\|Y^k_{t+h}-\hat Y^{\phi_t(k)}_{t+h}\|\le \epsilon^{\phi_t(k)}_{h}\bigr\}\Biggr)
\ge 1-\delta,
\end{equation}
which can be enforced by choosing confidence level $\overline\delta=\delta/(H\,|\Mcal_t|)$.
Then any feasible solution of \eqref{eq:mpc_optimization} satisfies
\begin{equation}
\mathbb P\Biggl(\,\bigcap_{k\in\Mcal_t}\;\bigcap_{h\in\mathcal{H}}
\{\,c(x_{t+h}, Y^k_{t+h})\ge 0\,\}\Biggr) \ge 1-\delta.
\end{equation}
\end{corollary}

\begin{proof}
Let $\mathcal E_t$ denote the event in \eqref{eq:scene_cov_event_heterogeneous}. Then the proof is analogous to~\propo{mpc_total_constraint_bound}.
\end{proof}

The overall procedure of the proposed framework is summarized in Alg.~\ref{alg:H-PRAP}.
\begin{algorithm}[h!]
\caption{Heterogeneous Predictor-based Risk-Aware Planning (H-PRAP)}
\label{alg:H-PRAP}
\begin{algorithmic}[1]
\Require $\mathcal{P} {\,=\,} \{P^\ell\}_{\ell=1}^{L}$, training and calibration dataset $D_{\text{train}}$ and $D_{\text{cal}}$, failure probability $\overline{\delta}$, thresholds $\{\theta_{\ell}\}_{\ell=0}^{L}$, planning and prediction horizon $T$ and $H$
  \For{$P^\ell\in\mathcal{P}$} \# \texttt{offline calculation}
  \State Train $P^\ell$ on $D_\text{train}$ 
    \State Calibrate $\{\epsilon^\ell_{h}\}_{h {\,\in\,\Hcal}}$ on $D_\text{cal}$ $\forall P^l {\,\in\,} \mathcal{P}$ under $\overline{\delta}$
    \State $(\sigma^\ell_h)^2\gets(\epsilon^\ell_{h})^2/\chi^2_n(1-\overline{\delta})$
  \EndFor
\For{$t$ from $0$ to $\infty$} \# \texttt{online planning loop}
  \State Observe $\xvect_t$ and $Y_t^k$
  \For{$k\in\mathcal{N}_t$}
    \State Compute $P^k_{t,\mathrm{col}}$ by~\eqref{eq:collision_probability_chi_squared} and~\eqref{eq:risk_aggregation_general}
    \State $\phi_t(k) \gets \max\{\ell{\,\in\,}\{0,\ldots,L\} : P^{k}_{t,\mathrm{col}} {\,\ge\,} \theta_\ell\}$
  \EndFor
  \State $\mathcal{M}_t\gets\{\,k\in\mathcal{N}_t:\phi_t(k)\neq0\,\}$
  \For{$k\in\mathcal{M}_t$}
    \State Obtain $\{\hat{Y}^k_{t+h}\}_{h\in{\mathcal{H}}}$ with $P^{\phi_t(k)}$ using $Y^k_{0:t}$
  \EndFor
  \State Obtain control inputs $\uvect_{t:t+T-1}$ as the solution of~\eqref{eq:mpc_optimization}
  \State Execute $\uvect_t$ to $f(\xvect_t, \uvect_t)$
\EndFor
\end{algorithmic}
\end{algorithm}

\section{Planning under Computational Constraints}\label{sec:H-PRAP_constraint}
The analysis in Section~\ref{sec:framework} assumed unlimited computational resources. However, safety-critical real-time systems must operate within a fixed compute time. 

First, we formalize this compute time constraint and analyze the resulting safety-efficiency trade-off.
Each control cycle has a hard deadline $T_{\text{total}}$. We reserve $T_{\text{mpc}}$ for solving the MPC in~\eqref{eq:mpc_optimization}, leaving a prediction budget
\begin{equation}
 B \triangleq T_{\text{total}} - T_{\text{mpc}} {\,>\,} 0.   
\end{equation}
Let $c^\ell{\,>\,}0$ denote the (approximately constant) per-obstacle runtime of predictor $P^\ell$, with higher-fidelity predictors being more expensive (e.g., $c^1 {\,\ge\,} c^2 {\,\ge\,} \cdots$). After the router $\phi_t$ assigns predictors to $\Mcal_t$, the requested time is defined as $T_{\mathrm{req}}(\phi_t){\,\triangleq\,}\sum_{k\in\Mcal_t} c^{\phi_t(k)}$.

When $T_{\mathrm{req}}(\phi_t) {\,\le\,} B$, all the assigned predictions are executed. Otherwise, H-PRAP employs \emph{risk-ordered dropping}, iteratively removing predictions with the smallest P-CRI until the budget is satisfied. This divides $\mathcal{M}_t$ into a \emph{enforced} set $\Mcal_t^{\mathrm{enf}}(B)$ (sent to predictors) and a \emph{dropped} set $\Mcal_t^{\mathrm{drop}}(B)$
who are excluded from predictions. 
Thus, if $k {\,\in\,} \Mcal_t^{\mathrm{enf}}(B)$ and $j {\,\in\,} \mathcal{M}_t^{\text{drop}}(B)$, then $P_{\mathrm{col}}^{k} {\,\ge\,} P_{\mathrm{col}}^{j}$.

A fixed budget entails a trade-off between \emph{quantity} (predicting many obstacles) and \emph{quality} (using tighter predictors), which respectively drive \emph{safety} and \emph{trajectory efficiency}.
Let
\begin{equation}
   P_{\text{res}}(B) \triangleq \sum_{k \in \Mcal_t^\text{drop}(B)} P_{\mathrm{col}}^k
\end{equation}
denote the \emph{residual risk}, i.e., the cumulative risk of obstacles that were sensed but not predicted due to the budget limit $B$.
In other words, $P_{\text{res}}(B)$ quantifies the portion of potential collision risk that remains unaccounted for under budget $B$.
By construction, the residual risk decreases monotonically with the budget: for any $0 {\,\le\,} B_1 {\,\le\,} B_2$,
\begin{equation}\label{eq:resiudal_risk}
   P_{\text{res}}(B_1) \,\geq\, P_{\text{res}}(B_2).
\end{equation}
This is trivial, since larger budgets require fewer removals
and $\Mcal_t^{\mathrm{drop}}(B_2) {\,\subseteq\,} \Mcal_t^{\mathrm{drop}}(B_1)$.

Second, we analyze how the choice of predictors impacts trajectory efficiency, measured by the number of time steps required for the agent to reach the goal. This analysis provides the theoretical basis for the efficiency results observed in our experiments.
For each step $s{\,\ge\,}0$, define the feasible set of state-control sequences that reach $\xvect_{s}{\,=\,} \xvect_g$:
\begin{equation}
\begin{aligned}
\mathcal{F}_s&\big(\{\epsilon_{h}^{\ell}\}_{h\in\mathcal{H}}\big) \triangleq
\Big\{\,(\xvect_{0:s},\uvect_{0:s-1})\ \Big|\ \\
&\xvect_{i+1} = f(\xvect_{i}, \uvect_{i}),\  \\
&\uvect_{i}\in\mathcal{U},\ \xvect_{i+1}\in\mathcal{X},\ \forall i=0,\ldots,s-1,\\
& c\big(\xvect_{s+h}, \hat{Y}_{s+h}^{\phi_s(k)}\big)\ {\,\ge\,} K\epsilon_{h}^{\phi_s(k)},\forall k{\,\in\,}\Mcal_s, \\
& \xvect_{s}=\xvect_g, \forall h {\,\in\,}1,\ldots,\min(H,s)\}\Big\}.
\end{aligned}
\end{equation}
We also define the earliest time at which the goal $\xvect_g$ is reachable by the feasible set of state-control sequence:
\begin{equation}
t_{\mathrm{arr}}\!\big(\{\epsilon_{h}^{\ell}\}\big)
=\min\big\{\,s:\ \mathcal{F}_s\!\big(\{\epsilon_{h}^{\ell}\}\big)\neq\varnothing\,\big\},
\end{equation}
where $t_{\mathrm{arr}} {\,=\,} \infty$ if $\mathcal{F}_s {\,=\,} \varnothing$ for all $s$.

\begin{proposition}[Impact of Conformal Radii on Steps-to-Goal]\label{prop:optimality_vs_radius}
Fix the predictor indices $\ell{\,\in\,}\{1,2\}$ for all $k{\,\in\,}\Mcal_t$, and hold $\{\hat{Y}^{\ell}_{t+h}(k)\}$ fixed for $h{\,\in\,}\Hcal$.
Given conformal radii $\{\epsilon^{\ell}_{h}\}_{h \in \mathcal{H}}$ and any nonnegative increments
$\{\Delta_{h}\}_{h\in\mathcal{H}}$, enlarge the radii element-wise to
$\epsilon^{\ell}_{h}+\Delta_{h}$.
Then
\begin{equation}
t_{\mathrm{arr}}\!\big(\{\epsilon^{\ell}_{h}{+}\Delta_{h}\}_{h\in\mathcal{H}}\big)\ \ge\ t_{\mathrm{arr}}\!\big(\{\epsilon^{\ell}_{h}\}_{h\in\mathcal{H}}\big),
\label{eq:tarr_monotone}
\end{equation}
where $t_{\mathrm{arr}}\!\big(\{\epsilon^{\ell}_{h}\}_{h\in\mathcal{H}}\big)$ is the earliest time step after $t$ at which the goal $\xvect_g$
is reached by \emph{some} trajectory satisfying the MPC constraints in~\eqref{eq:mpc_optimization}
(i.e., \eqref{eq:mpc_dynamics}-\eqref{eq:mpc_safety_runningEx}); if no such time exists, set $t_{\mathrm{arr}}{\,=\,}+\infty$.
\end{proposition}

\begin{proof}
Enlarge the conformal radii element-wise to $\epsilon^{\ell}_{h}{\,+\,}\Delta_{h}$ with $\Delta_{h}{\,\ge\,}0$.
If we take any $(\xvect_{0:s},\uvect_{0:s-1})\in \mathcal{F}_s\!\big(\{\epsilon_h^{\ell}{+}\Delta_h\}_{h\in\mathcal{H}}\big)$,
for every $k{\,\in\,}\Mcal_t$ and $h {\,\in\,} \mathcal{H}$,
\begin{equation*}
\begin{aligned}
c\big(\xvect_{t+h},\hat{Y}^{\ell}_{t+h}(k)\big)
\ge K\big(\epsilon^{\ell}_{h}{+}\Delta_{h}\big) \ge K\,\epsilon^{\ell}_{h},
\end{aligned}
\end{equation*}
so the same sequence also satisfies \eqref{eq:mpc_safety_runningEx} under $\{\epsilon_{h}^{\ell}\}$.
Because \eqref{eq:mpc_dynamics} and \eqref{eq:mpc_constraints} are unchanged, it follows that
\begin{equation*}
\mathcal{F}_s\!\big(\{\epsilon_{h}^{\ell}{+}\Delta_{h}\}_{h\in\mathcal{H}}\big)\ \subseteq\ \mathcal{F}_s\!\big(\{\epsilon_{h}^{\ell}\}_{h\in\mathcal{H}}\big),
\qquad\forall\,s\ge0.
\end{equation*}
Taking $\min\{s:\cdot\}$ over a subset cannot yield a smaller value than taking it over its superset (with $+\infty$ for emptiness). Hence
\begin{equation*}
\begin{aligned}
t_{\mathrm{arr}}\!\big(\{\epsilon_{h}^{\ell}{+}\Delta_{h}\}_{h\in\mathcal{H}}\big)
&= \min\big\{\,t{\,+\,}s{\,:\,} \mathcal{F}_s\!\big(\{\epsilon_{h}^{\ell}{+}\Delta_{h}\}_{h\in\mathcal{H}}\big){\,\neq\,}\varnothing\big\}\\
&\ge \min\big\{\,t{\,+\,}s{\,:\,}\mathcal{F}_s\!\big(\{\epsilon_{h}^{\ell}\}_{h\in\mathcal{H}}\big){\,\neq\,}\varnothing\big\} \\
& = t_{\mathrm{arr}}\!\big(\{\epsilon_{h}^{\ell}\}_{h\in\mathcal{H}}\big).\mkern130mu\qedhere
\end{aligned}
\end{equation*}
\end{proof}
In numerical experiments, we report the number of steps taken by the MPC until it reaches $\xvect_g$. This quantity is lower-bounded by $t_{\mathrm{arr}}(\cdot)$. Hence, enlarging the conformal radii cannot reduce the reported steps-to-goal, consistent with Proposition~\ref{prop:optimality_vs_radius}.
Moreover, the results of Proposition~\ref{prop:optimality_vs_radius} can be readily extended to more than two predictors.

Together, \eqref{eq:resiudal_risk} and \eqref{eq:tarr_monotone} formalize the core trade-offs.
The former shows that larger budgets reduce unprotected risk mass, while the latter explains why low-fidelity predictions degrade trajectory efficiency. H-PRAP navigates this trade-off by allocating high fidelity only where P-CRI indicates it most improves safety, while keeping lightweight predictions elsewhere to satisfy the budget and mitigate performance loss in planning.
\begin{figure}
    \centering
    \includegraphics[width=\linewidth]{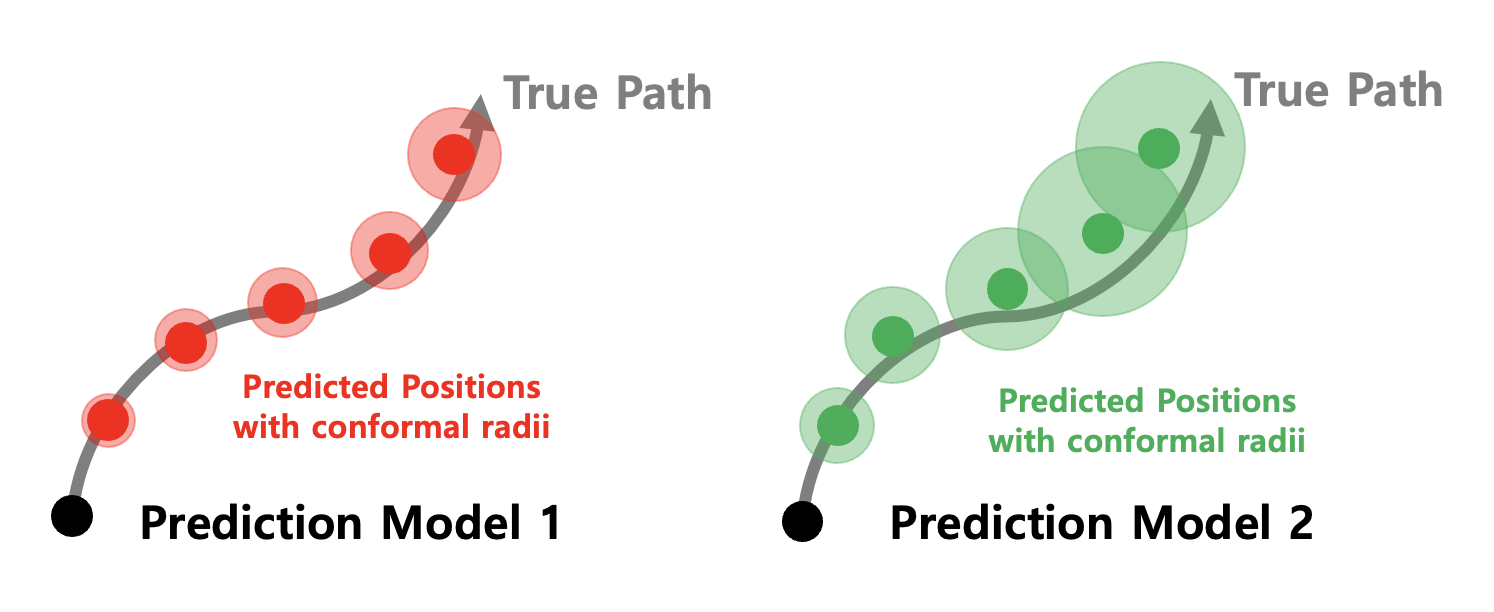}
    \caption{Sample visualization of $P^1$ versus $P^2$. Since $P^1$ is more accurate than $P^2$, the circular regions (shaded) around each prediction point (solid dot) are smaller compared to those of $P^2$.} \label{fig:Pred1_vs_Pred2}
    \vspace{-1.5em}
\end{figure}

\section{Numerical Experiments}\label{sec:experiments}

\subsection{Setting}
\begin{figure}[t]
    \centering
    \includegraphics[width=0.95\linewidth]{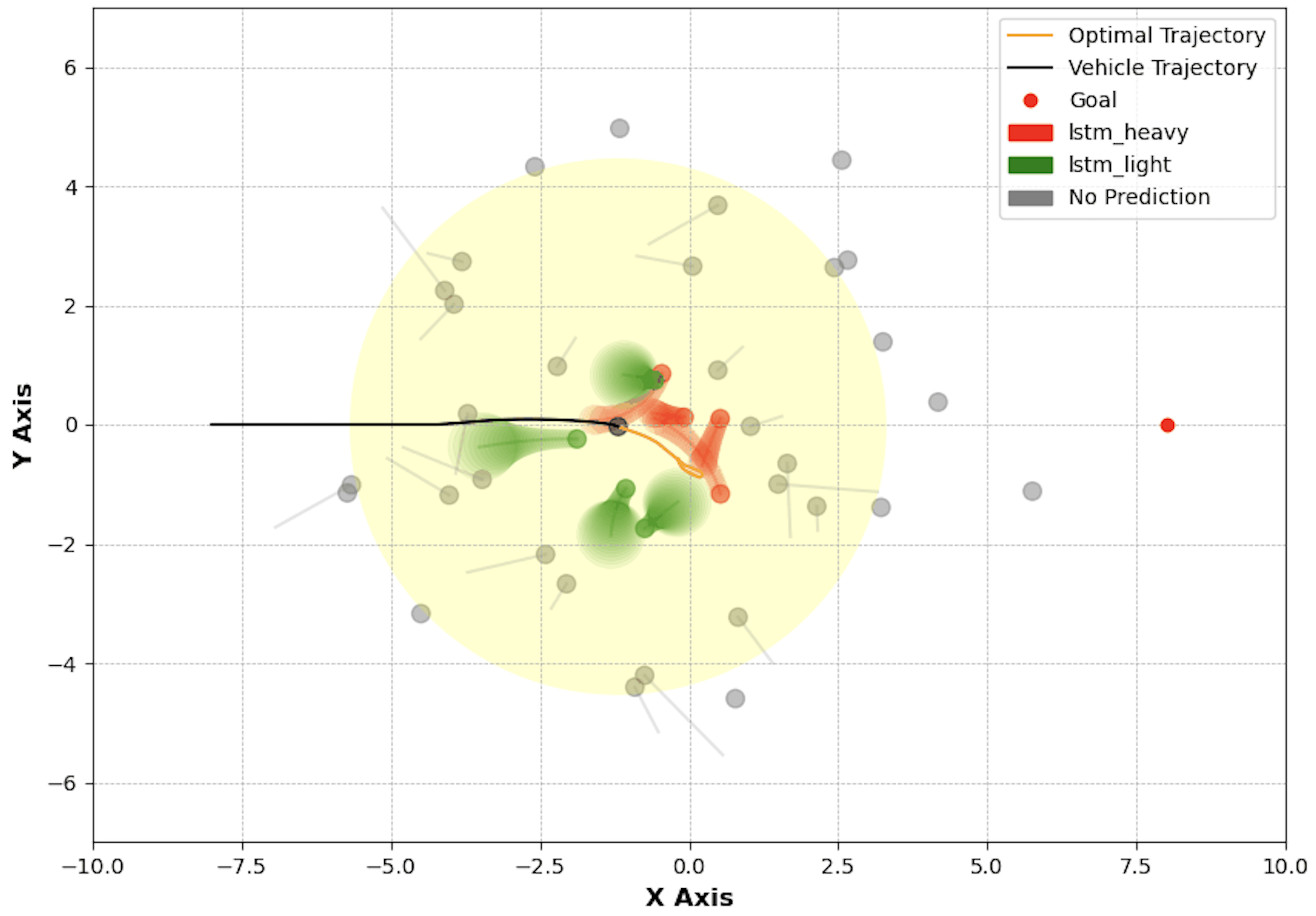}
    \caption{Example screenshot of the agent navigating with H-PRAP.}
    \label{fig:simulation_40}
    \vspace{-1.5em}
\end{figure}

Here, we compare H-PRAP against single-predictor baselines under prediction compute budget.
We employ two LSTM predictors to span the accuracy--efficiency spectrum: a higher-fidelity, higher-cost predictor $P^1$ and a lower-fidelity, lower-cost predictor $P^2$. Both are LSTM models with different number of parameters (2-layer, 128 hidden-nodes and single-layer, 32 hidden-nodes, respectively). H-PRAP routes obstacles to $P^1$ or $P^2$ based on P-CRI. 
The agent is modeled as a kinematic differential drive mobile robot (DDMR)~\cite{siegwart2011introduction} as follows:
\begin{equation}
\begin{bmatrix}
x_{t+1} \\ y_{t+1} \\ \theta_{t+1}
\end{bmatrix}
=
\begin{bmatrix}
    x_t \\ y_t \\ \theta_t
\end{bmatrix}
+
\Delta t
\begin{bmatrix}
\cos(\theta_t) & 0 \\
\sin(\theta_t) & 0 \\
0 & 1
\end{bmatrix}
\begin{bmatrix}
    v_t \\ w_t
\end{bmatrix},
\end{equation}
where $v_t$ and $w_t$ denote the linear and angular velocities of the agent, respectively.
The MPC problem~\eqref{eq:mpc_optimization} is posed and solved using IPOPT solver.
For the planning configuration, planning and prediction horizon are set to $T{\,=\,}H{\,=\,}30$. The total compute time for each cycle is $T_{\text{total}} {\,=\,} T_{\text{mpc}} {\,+\,} B$, where $T_{\text{mpc}} {\,=\,} 100\,\text{ms}$ is the minimum guaranteed time reserved for solving MPC. We sweep the budget $B {\,\in\,} \{ 30,35,40,45,50,55,60 \}\, \text{ms}$ and run 500 randomized episodes containing $40$ dynamic obstacles which are randomly initialized with different motion patterns. The failure rate is set to $\delta=0.05$ (confidence level $1-\delta=0.95$) to calibrate the conformal radius. Safety is evaluated by the \emph{success rate} [\%], defined as the fraction of episodes where the agent reaches the goal without any collision. 
Efficiency is evaluated by the \emph{number of steps to the goal} in the successful episodes.
We compare three baselines:
\begin{itemize}
    \item \textbf{SP1 (Heavy only):} single-predictor architecture using $P^1$ (prioritizing prediction quality).
    \item \textbf{SP2 (Light only):} single-predictor architecture using $P^2$ (prioritizing prediction quantity).
    \item \textbf{H-PRAP:} heterogeneous predictor architecture using $P^1$ and $P^2$ routed by P-CRI.
\end{itemize}
Fig.~\ref{fig:Pred1_vs_Pred2} visualizes the comparison of regions bounded by conformal prediction between two predictors $P^1$ and $P^2$, and Fig.~\ref{fig:simulation_40} shows the agent navigating through dense dynamic obstacles leveraging H-PRAP.

\subsection{Results and Analysis}\label{sec:results}

The experimental results are best understood through a prediction compute-budget--induced trade-off between obstacle prediction \emph{quantity} (driving safety) and prediction \emph{quality} (driving trajectory efficiency). 
\eqref{eq:resiudal_risk} and \eqref{eq:tarr_monotone} formalize these two sides. We now analyze how each architecture resolves this trade-off, with results shown in Table~\ref{tab:performance}.

\noindent\textbf{SP2 (Quantity over Quality).}
SP2 adopts a quantity-first strategy. Its fast, low-fidelity predictor keeps the dropped set small, inducing little to no residual risk $P_\text{res}(B)$. This allows it to maintain a high success rate even under tight budgets. However, this safety comes at the cost of trajectory efficiency. Larger conformal radii $\{\epsilon_h^2\}_{h\in\mathcal{H}}$ degrade trajectory efficiency, as reflected by increased steps-to-goal.
Moreover, these larger radii shrink the MPC feasible set for certain scenarios, which can sometimes induce solver infeasibility (deadlocks), slightly reducing its success rate despite sufficient budget.

\noindent\textbf{SP1 (Quality over Quantity).}
SP1 is a quality-first architecture. When the budget is sufficient and all obstacles are enforced, its high-fidelity predictor with smaller radii yields shorter steps-to-goal and high trajectory efficiency, as implied by Proposition~\ref{prop:optimality_vs_radius}. The weakness of this strategy is exposed when the budget is tight. Because each prediction is costly, SP1 can cover only a limited number of obstacles, which results in extensive risk-ordered dropping. As a result, the dropped set $\mathcal{M}_t^{\text{drop}}(B)$ grows, the residual risk $P_{\text{res}}(B)$ increases, and safety is substantially degraded.

\noindent\textbf{H-PRAP (Balancing Safety and Efficiency).}
H-PRAP uses P-CRI to balance the trade-off between the two extremes described above. It allocates expensive prediction only to high-risk obstacles and lightweight prediction elsewhere, jointly maintaining quantity and quality.

When the budget is \emph{tight}, its key advantage emerges. It allocates the heavy predictor judiciously dropping far fewer obstacles than SP1 and thus suffering much less safety degradation.
When the budget is \emph{sufficient}, fewer obstacles are dropped. Unconstrained by the budget, H-PRAP's P-CRI routing prioritizes the high-fidelity predictor for all high-risk obstacles to optimize the trajectory, thus achieving trajectory efficiency comparable to SP1.

The average steps-to-goal decreases as the budget tightens. This is largely a selection effect. Since we report steps only on successful episodes, tighter budgets filter out many difficult scenarios (which typically require detours or extensive avoidance, and thus incur larger steps), thereby lowering steps-to-goal.
%

Overall, H-PRAP attains a favorable point in the safety-efficiency trade-off across a wide range of budgets, demonstrating robustness for real-time planning. By strategically assigning high-fidelity prediction to high-risk obstacles and lightweight prediction elsewhere, H-PRAP achieves the \textit{safety of SP2} and the \textit{trajectory efficiency of SP1}, avoiding both the drop-induced safety loss of SP1 and the inherent path inefficiency of SP2.

\begin{table}[t]
    \centering
    \begin{tabular}{c|ccc|ccc}
        \toprule
        \multirow{2}{*}{Budget $B$} & \multicolumn{3}{c|}{Success Rate [\%]} & \multicolumn{3}{c}{Avg. Steps-to-Goal}\\
        \cmidrule(lr){2-4} \cmidrule(lr){5-7}
        [ms] & H-PRAP & SP1 & SP2 & H-PRAP & SP1 & SP2 \\
        \midrule
        30 & 93.0 & 71.0 & \textbf{93.8} & 167.3 & \textbf{119.7} & 193.2 \\
        35 & \textbf{94.8} & 83.8 & 94.0 & 171.2 & \textbf{144.0} & 192.8 \\
        40 & \textbf{95.8} & 90.8 & 94.0 & 172.3 & \textbf{156.5} & 192.3 \\
        45 & \textbf{96.0} & 94.2 & 94.4 & 173.5 & \textbf{164.2} & 192.7 \\
        50 & \textbf{96.0} & 94.8 & 95.4 & 173.5 & \textbf{165.3} & 194.2 \\
        55 & \textbf{96.4} & \textbf{96.4} & 95.4 & 173.8 & \textbf{167.9} & 194.1 \\
        60 & 97 & \textbf{97.4} & 95.8 & 174.5 & \textbf{169.9} & 194.9 \\
        \bottomrule
    \end{tabular}
    \caption{Performance comparison of H-PRAP, SP1, and SP2 under different compute budget $B$.}
    \label{tab:performance}
    \vspace{-1.5em}
\end{table}


\section{Conclusion}

We proposed H-PRAP, a heterogeneous predictor-based framework for safe and efficient planning in dense, uncertain environments. The framework uses our novel Probability-based Collision Risk Index (P-CRI) to dynamically allocate computational effort to high-risk obstacles.
We analyzed that H-PRAP achieves the best balance of safety and efficiency of planning trajectory when budget is restricted, compared to architectures which only use a single predictor.
Numerical experiments validated our analysis under various scenarios, demonstrating that H-PRAP best balances between safety and trajectory efficiency than other methods under limited compute budget.
Future work will focus on refining P-CRI by estimating the probability of active constraints at the MPC solver level, incorporating interaction-aware predictors, and extending the framework to multi-agent planning.


\bibliographystyle{IEEEtran}
\bibliography{refs}

\end{document}